\documentclass[11pt,a4paper]{article}
\usepackage[hyperref]{acl2017}
\usepackage{times,amsmath}
\usepackage{latexsym,bbm}
\usepackage{todonotes}
\usepackage{booktabs}
\usepackage{multirow}
\usepackage{etoolbox}

\usepackage{url}

\usepackage[algo2e,ruled,vlined,linesnumbered]{algorithm2e}

\newlength{\chartheight}
\aclfinalcopy 
\def\aclpaperid{31} 

\title{Reward-Balancing for Statistical Spoken Dialogue Systems using Multi-objective Reinforcement Learning}

\author{Stefan~Ultes, Pawe\l~Budzianowski, I\~nigo~Casanueva, Nikola~Mrk{\v s}i{\' c},\\ { \bf Lina~Rojas-Barahona, Pei-Hao~Su, Tsung-Hsien~Wen, Milica~Ga{\v s}i{\' c} and Steve~Young} \\
  Engineering Department, University of Cambridge, Cambridge, United Kingdom \\
  {\tt \{su259,pfb30,ic340,nm480,lmr46,phs26,thw28,mg436,sjy11\}@cam.ac.uk}}

\date{24.04.2017}

\begin{document}
\maketitle
\begin{abstract}
Reinforcement learning is widely used for dialogue policy optimization where the reward function often consists of more than one component, e.g., the dialogue success and the dialogue length. In this work, we propose a structured method for finding a good balance between these components by searching 
for the optimal reward component weighting.
To render this search feasible, we use multi-objective reinforcement learning to significantly reduce the number of training dialogues required. We apply our proposed method to find optimized component weights for six domains and compare them to a default baseline.
\end{abstract}

\section{Introduction}

In a Spoken Dialogue System (SDS), one of the main problems is to find appropriate system behaviour for any given situation. This problem is often modelled using reinforcement learning (RL) where the task is to find an optimal policy $\pi(b) = a$ which maps the current belief state $b$---an estimate of the user goal---
to the next system action $a$. To do this, RL algorithms seek to optimize an objective function, the reward $r$, using 
sample dialogues. In contrast to other RL tasks (like AlphaGo~\cite{silver2016mastering}), the reward used in goal-oriented dialogue systems usually consists of more than one objective (e.g.,  task success and  dialogue length~\cite{levin1998b,lemon2006,young2013}). 

However, balancing these rewards is rarely considered and the goal of this paper is to propose a structured method for finding the optimal weights for a multiple objective reward function. Finding a good balance between multiple objectives is usually domain-specific and not straight-forward. For example, in the case of task success and dialogue length, if the  reward for success is too high, the learning algorithm is insensitive to potentially irritating actions such as {\tt repeat} provided that the dialogue is ultimately successful.
Conversely, if the reward for success is too small, the resulting policy may irritate users by offering inappropriate solutions before fully illiciting the user's requirements.

In this paper, we propose to find a suitable reward balance by searching through the space of reward component weights. 
Doing this with conventional RL techniques is infeasible as a policy must be trained for each candidate balance and this requires an enormous number of training dialogues. To alleviate this, we propose to use multi-objective RL (MORL) which is specifically designed for this task (among others~\cite{roijers2013survey}). 
Then, only one policy needs to be trained which may be evaluated with several candidate balances. 
To the best of our knowledge, this is the first time MORL has been applied to dialogue policy optimization.

In contrast to previous work which explicitly selects component weights to maximize user satisfaction~\cite{walker2000} explicitly, 
the proposed method enables optimisation of an implicit goal by allowing the interplay each reward component to be explored at low computational cost.

Several different algorithms have previously been used for MORL~\cite{castelletti2013multiobjective,van2015risk,pirotta2015multi,DBLP:journals/corr/MossalamARW16}. In this work, we propose a novel MORL algorithm based on Gaussian processes.  This is described in Section~\ref{sec:morl} along with a brief introduction to MORL. In Section~\ref{sec:rewardbalancing}, the proposed method for finding a good reward balance with MORL is presented. Section~\ref{sec:experiments} describes the application and evaluation of the balancing method on six different domains.  Finally conclusions are drawn in Section~\ref{sec:conclusion}.


\section{Multi-objective Reinforcement Learning with Gaussian Processes}
\label{sec:morl}
In this Section we present our proposed extension of the GPSARSA algorithm for MORL after giving a brief introduction to single- and multi-objective RL and the GPSARSA algorithm itself.

\paragraph{Reinforcement Learning}
Reinforcement learning (RL) is used in a sequential decision-making process where a decision-model (the policy $\pi$) is trained based on sample data and a potentially delayed objective signal (the reward $r$)~\cite{sutton1998}. Implementing the Markov assumption, the policy selects the next action $a \in A$ based on the current system belief state $b$ to optimise the accumulated future reward $R_t$ at time $t$:
\begin{equation}
\label{eq:acR}
R_t = \sum_{k=0}^{\infty} \gamma^k r_{t+k+1} \; .
\end{equation}
Here, $k$ denotes the number of future steps, $\gamma$ a discount factor and $r_\tau$ the reward at time $\tau$.

The $Q$-function models the expected accumulated future reward $R_t$ when taking action $a$  in belief state $b$ and then following policy $\pi$:
\begin{equation}
Q^{\pi}(b,a) = E_{\pi}[R_t | b_t = b, a_t = a] \; .
\end{equation}

\paragraph{GPSARSA}
For most real-world problems, finding the exact optimal $Q$-values is not feasible. Instead, \citet{engel2005reinforcement} have proposed the GPSARSA algorithm which uses Gaussian processes (GP) to approximate the $Q$-function. \citet{gasic2014gaussian} have shown that this works well when applied to the problem of spoken dialogue policy optimisation. GPSARSA is a Bayesian on-line learning algorithm which models the $Q$-function as a zero-mean GP which is fully defined by a mean and a kernel function $k$: 
\begin{equation}
Q^{\pi}(b,a) \sim \mathcal{GP}(0,k(b,a),(b,a))) \; ,
\end{equation}
where the kernel models the correlation between data points. Based on sample data, the GP is trained to approximate $Q$ such that the variance derived from the kernel represents the uncertainty of the approximation.

In dialogue management, the following kernel has been successfully used: 
\begin{equation}
k((b,a),(b',a')) = \delta(a,a') \cdot k_{lin}(b,b') \; .
\end{equation}
It consists of a linear kernel for the continuous belief representation $b$ and the $\delta$-kernel for the discrete system action $a$.

\paragraph{Multi-objective Reinforcement Learning}
In multi-objective reinforcement learning (MORL), the objective function does not consist of only one but of many dimensions. Thus, the reward $r_t$ becomes a vector $\mathbf{r}_t = (r_t^1, r_t^2, \ldots, r_t^m)$, where $m$ is the number of objectives. 

To define the contribution of each objective, a scalarization function $f$ is introduced which uses weights $\mathbf{w}$ for the different objectives to map the vector representation to a scalar value. The solution to a MORL problem is a set of optimal policies containing an optimal policy for any given weight configuration.

In MORL, the $Q$-function may either be modelled as a vector of $Q$-functions 
or directly as the expectation of the scalarized vector of $(R_t^1 \ldots R_t^m)$: 
\begin{equation}
Q_\mathbf{w}^{\pi}(b) = E[f(\mathbf{R}_t, \mathbf{w}) | \pi, b, a] \; .
\label{eq:scalarQ}
\end{equation}
In practice, the scalarization function is often modelled as a linear function (the weighted sum):
\begin{equation}
f(\mathbf{r}_t,\mathbf{w}) = \sum_m w_m r_t^m \; .
\label{eq:weightedsum}
\end{equation}

\paragraph{Multi-objective GPSARSA}
The proposed multi-objective (MO) GPSARSA is based on Equation~\ref{eq:scalarQ}. By approximating the scalarized $Q$-function directly using a GP, the GPSARSA algorithm 
may be applied for MORL. The GP (and thus the $Q$-function) is extended by one parameter---the weight vector $\mathbf{w}$: $Q(b,a,\mathbf{w})$.

Approximating the $Q$-function with a GP relies on the fact that the accumulated future reward $R_t$ (Eq.~\ref{eq:acR}) may be decomposed as
\begin{equation}
R_t = r_{t+1} + \gamma R_{t+1} \; .
\end{equation}
Accordingly, for using a GP to directly estimate the scalarized reward in  MO-GPSARSA, the equation
\begin{align}
f(\mathbf{R}_{t}, \mathbf{w}) &= f(\mathbf{r}_{t+1} + \gamma \mathbf{R}_{t+1}, \mathbf{w}) \nonumber \\ 
&\overset{!}{=} f(\mathbf{r}_{t+1},\mathbf{w}) + \gamma f( \mathbf{R}_{t+1}, \mathbf{w}) 
\end{align}
must hold. This is true in case of using a linear scalarization function $f$ (Eq.~\ref{eq:weightedsum}).

To alter the kernel accordingly, a linear kernel for $\mathbf{w}$ is added to the state kernel\footnote{A similar type of kernel extension has been proposed previously in a different context, e.g., \cite{casanueva2015}.} resulting in
\begin{align}
k&((b,a,\mathbf{w}),(b',a',\mathbf{w}')) \nonumber \\ 
&= \delta(a,a') \cdot \left(k_{lin}(b,b') + k_{lin}(\mathbf{w},\mathbf{w}')\right) \; .
\end{align}
Since a linear scalarization function is applied, the correlations with other data points are also assumed to be linear.

To train a policy using multi-objective GPSARSA, a new weight configuration is sampled randomly for each training dialogue. An example of the training process being applied to dialogue policy optimization with the two objectives task success and dialogue length is depicted 
in Algorithm~\ref{alg:mogpsarsa}.

\begin{algorithm2e}[t]
\NoCaptionOfAlgo
\footnotesize
\DontPrintSemicolon
\KwIn{ dialogue success reward $r_s$,  dialogue length penalty $r_l$}


\ForEach{training dialogue}{
	select $w_s, w_l$ randomly \;
    execute dialogue and record $(b_t,a_t,\mathbf{w})$ in $D$ for each turn $t$\;
    \tcp{dialogue length penalty}
    $r \leftarrow w_l \cdot |D| \cdot r_l$ \;
    \tcp{dialogue success reward}
    \uIf{dialogue successful}{
        $r \leftarrow r + w_r \cdot r_s$
    }
    update GP using $D$ and $r$ \;
    reset $D$ \;
}

\caption{Algorithm~\ref{alg:mogpsarsa}: Training of the MO-GPSARSA.}

\label{alg:mogpsarsa}
\end{algorithm2e}








\section{Reward Balancing using MORL}
\label{sec:rewardbalancing}


The main contribution of this paper is to provide a structured method for finding a good balance between multiple rewards for learning dialogue policies. For the two-objective problem of having a task success reward $r_s$ and a dialogue length reward $r_l$, $\mathbf{r} = (r_s,r_l)$, the scalarized reward is
\begin{align}
r =  f(\mathbf{r},\mathbf{w}) &=  \mathbbm{1}_{TS} \cdot w_s r_s + T \cdot w_s r_l  \nonumber \\
 &= \mathbbm{1}_{TS} \cdot r_s^w + T \cdot r_l^w \; ,
\end{align}
where $T$ is the number of turns and $\mathbbm{1}_{TS} = 1$ iff the dialogue is successful, zero otherwise.

To find a good reward balance, we adopt the following procedure:
\begin{enumerate}
\item Set initial reward values $r_s^w$ and $r_l^w$ along with the initial weight configuration.
\item Apply MORL to train a policy for a given number of training dialogues and evaluate with different weight configurations.
\item Select an appropriate balance based on success-weight and length-weight curves to optimise the individual implicit goal.
\end{enumerate}

The method may be refined by applying it recursively with different grid sizes. After selecting a suitable weight configuration, a single-objective policy may be trained.




\section{Experiments and Results}
\label{sec:experiments}

\begin{figure*}[t]
  \centering
  \begin{minipage}[b]{0.33\linewidth}
    \centering
    \includegraphics[width=\linewidth,height=\chartheight]{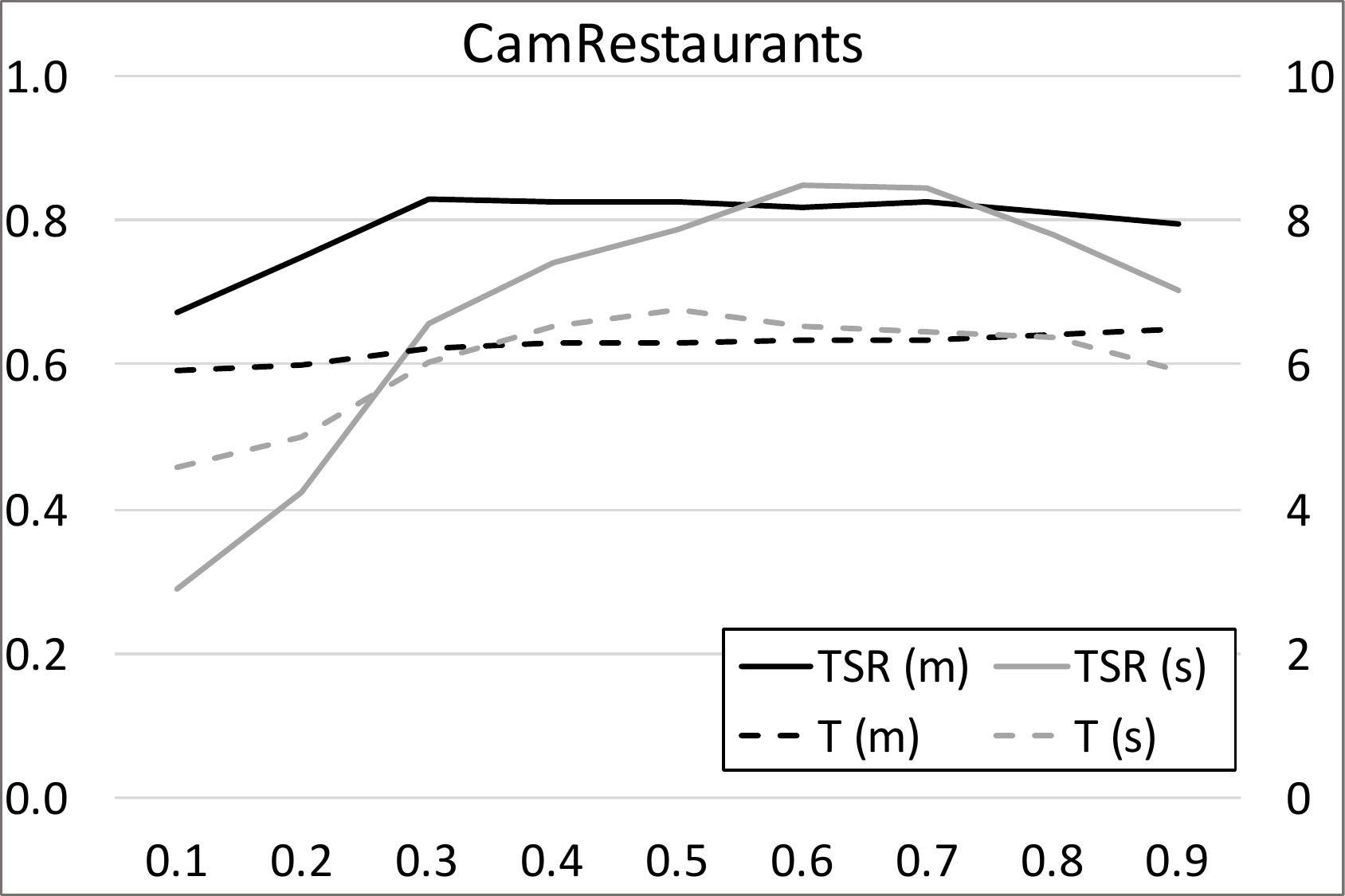}
  \end{minipage}%
  \hfill
  \begin{minipage}[b]{0.33\linewidth}
    \centering
    \includegraphics[width=\linewidth,height=\chartheight]{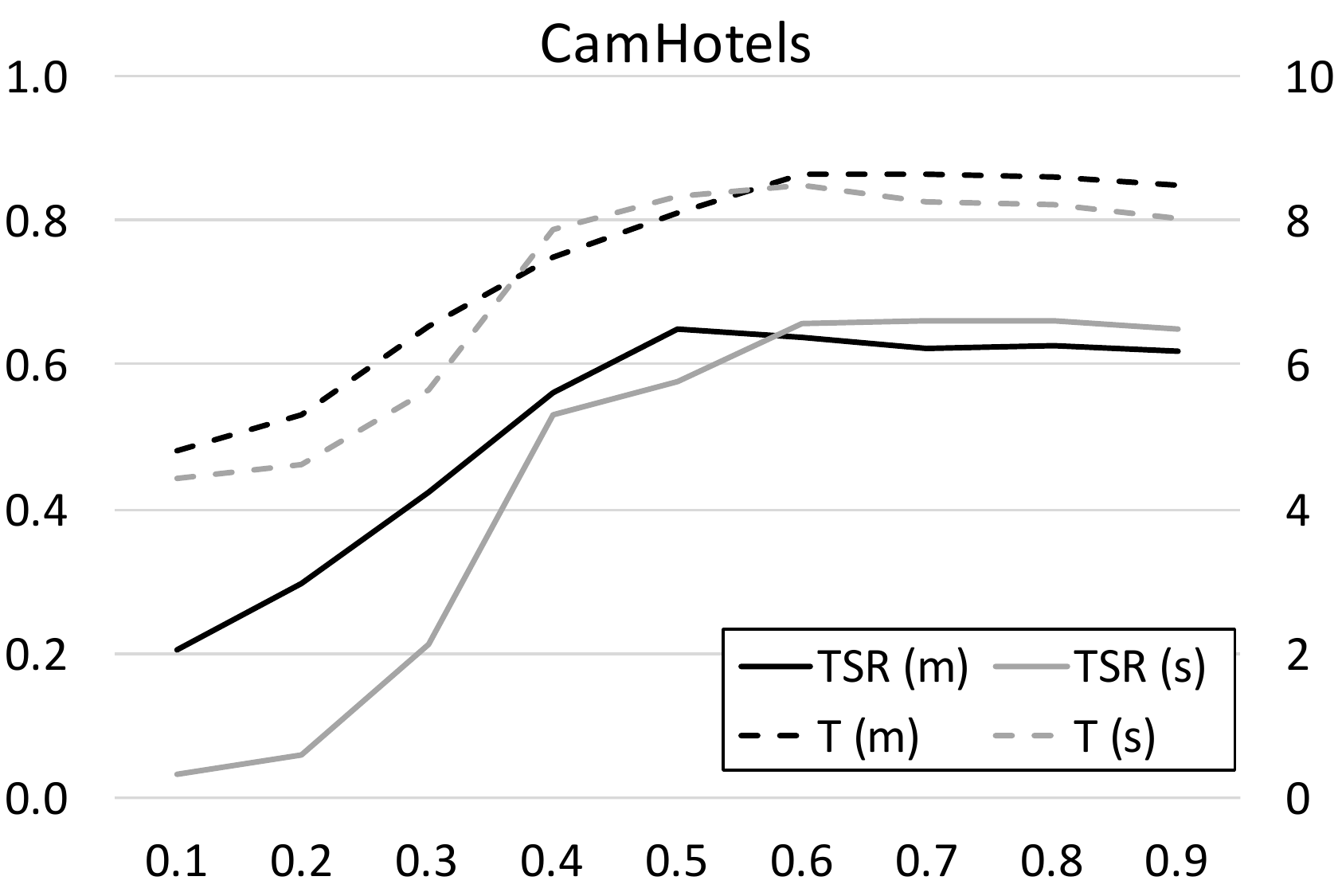}
  \end{minipage}%
  \hfill
  \begin{minipage}[b]{0.33\linewidth}
    \centering
    \includegraphics[width=\linewidth,height=\chartheight]{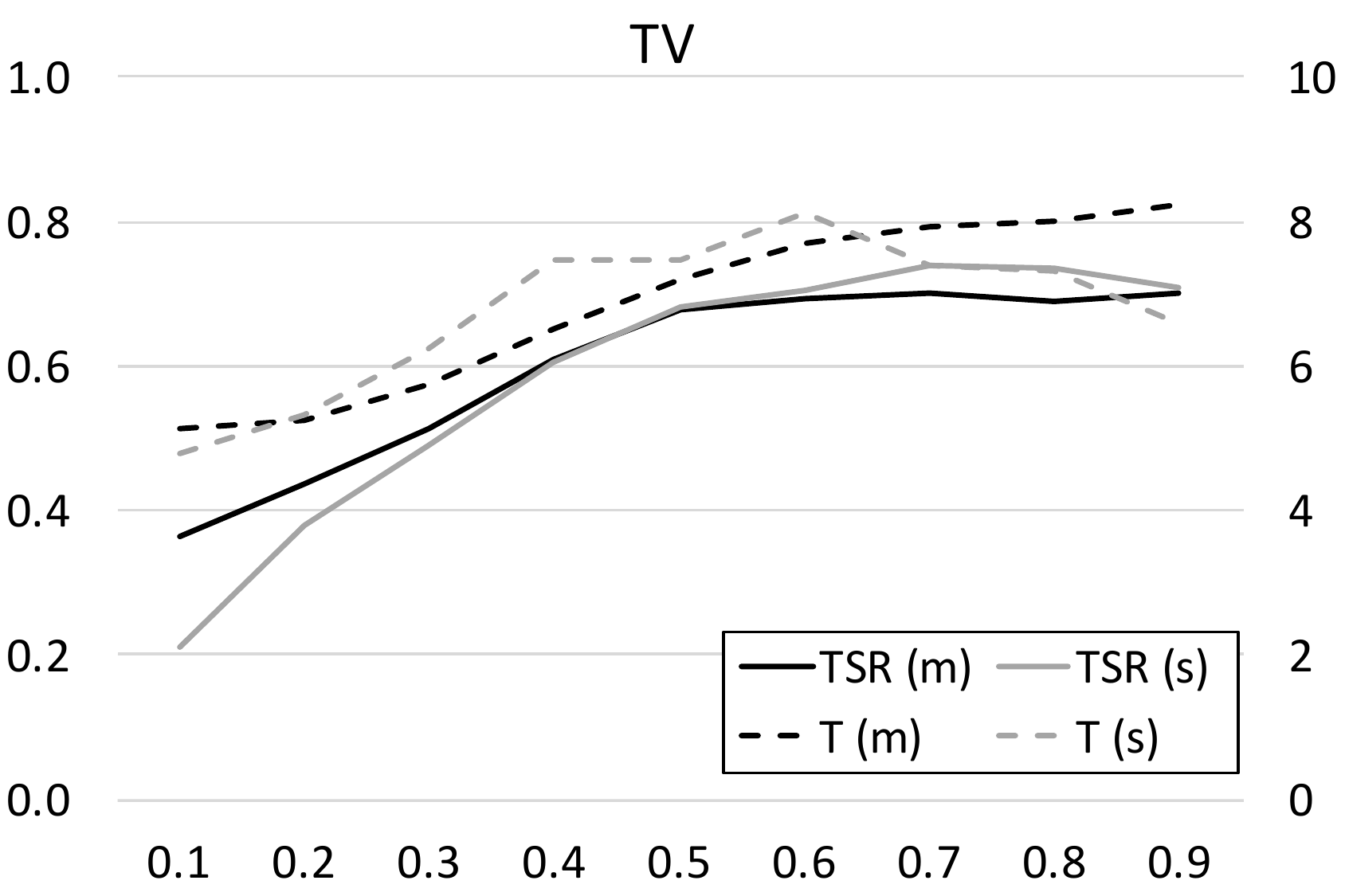}
  \end{minipage}%
  
  \begin{minipage}[b]{.33\linewidth}
    \centering
    \includegraphics[width=\linewidth,height=\chartheight]{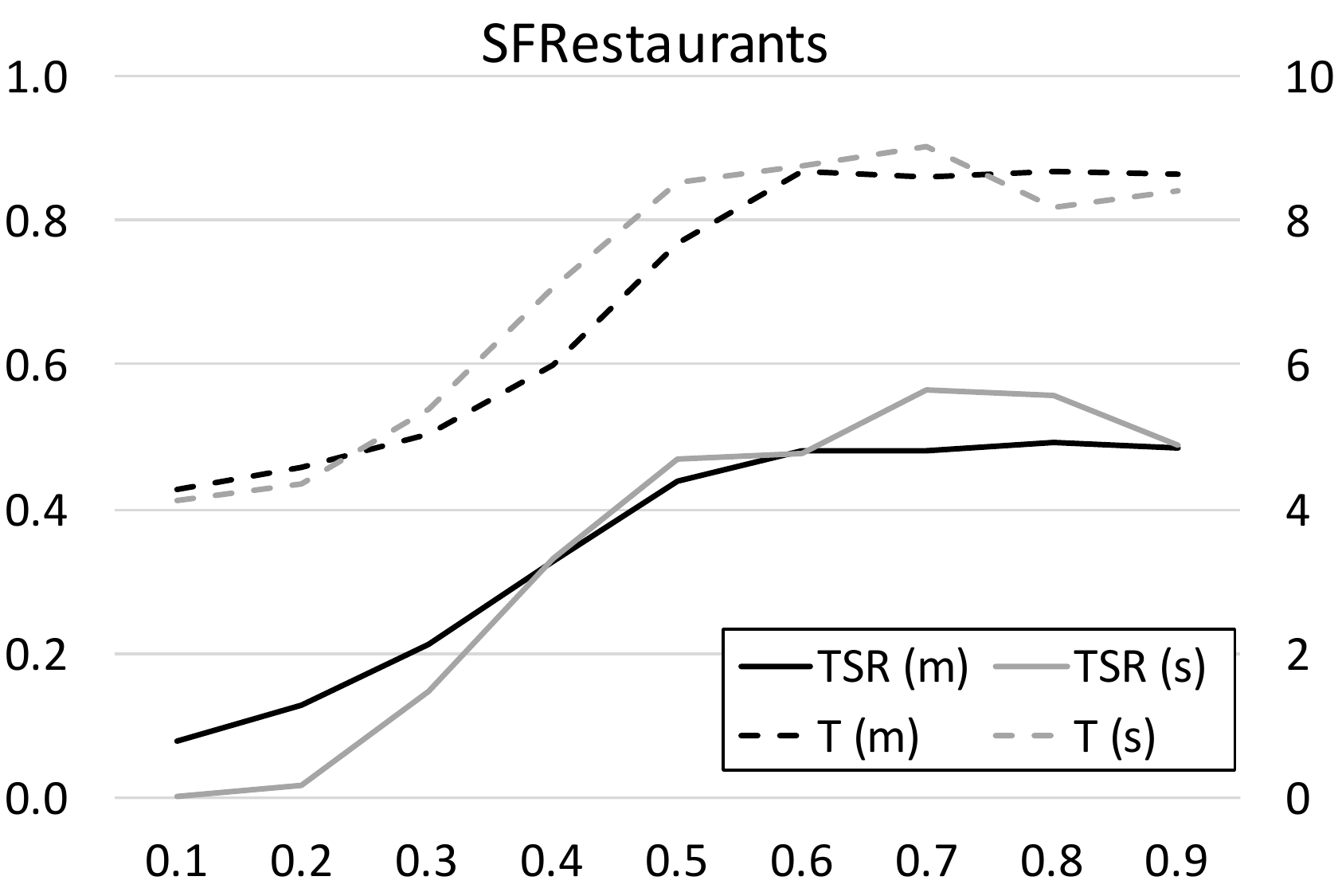}
  \end{minipage}%
  \hfill
  \begin{minipage}[b]{0.33\linewidth}
    \centering
    \includegraphics[width=\linewidth,height=\chartheight]{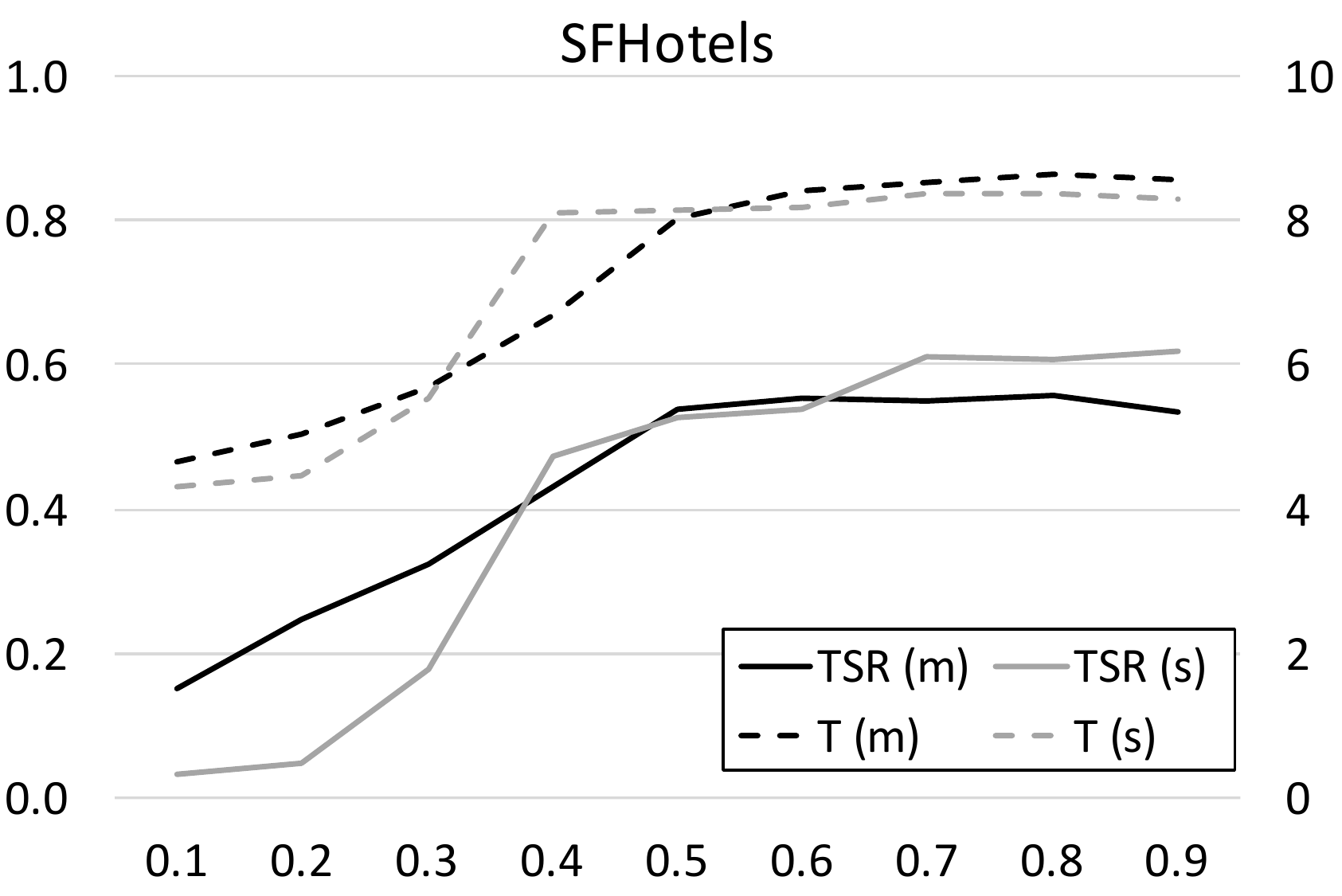}
  \end{minipage}%
  \hfill
    \begin{minipage}[b]{.33\linewidth} 
    \centering
    \includegraphics[width=\linewidth,height=\chartheight]{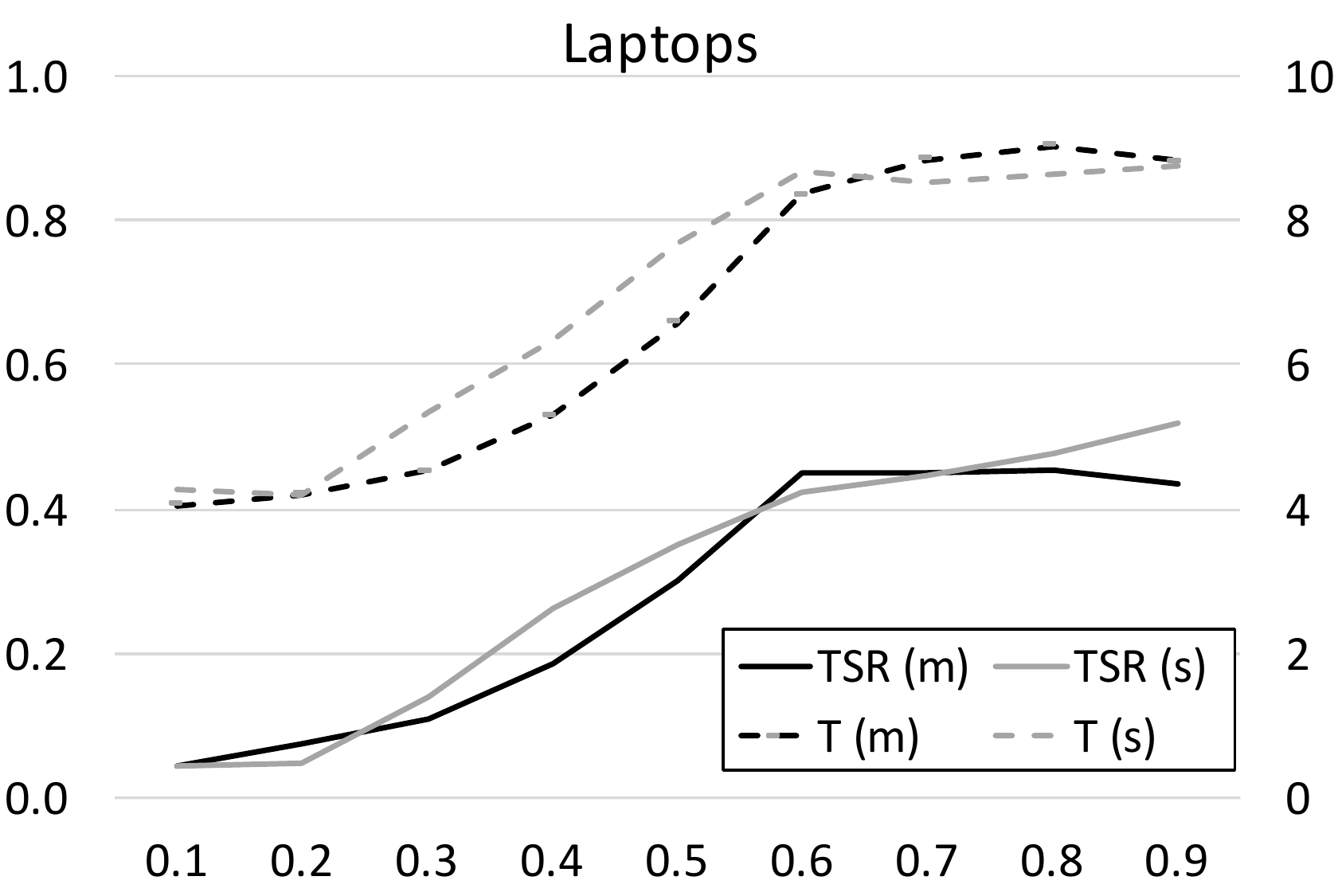}
  \end{minipage}%
  \caption{The MORL success-weight and length-weight curves (\textit{m}, task success rate (TSR) on left, number of turns T on right vertical axes; success weights $w_s$ on horizontal axes) after 3,000 training dialogues. Each data point is the average over five policies with different seeds where each policy/weight configuration is evaluated with 300 dialogues. As a comparison, the same curves using single-objective RL (\textit{s}, separate policies trained for each balance) have been created \textit{after} selecting the weights.}
  \label{fig:weight_curves}
\end{figure*}

\begin{figure*}[t]
  \centering
  \begin{minipage}[b]{0.33\linewidth}
    \centering
    \includegraphics[width=\linewidth,height=\chartheight]{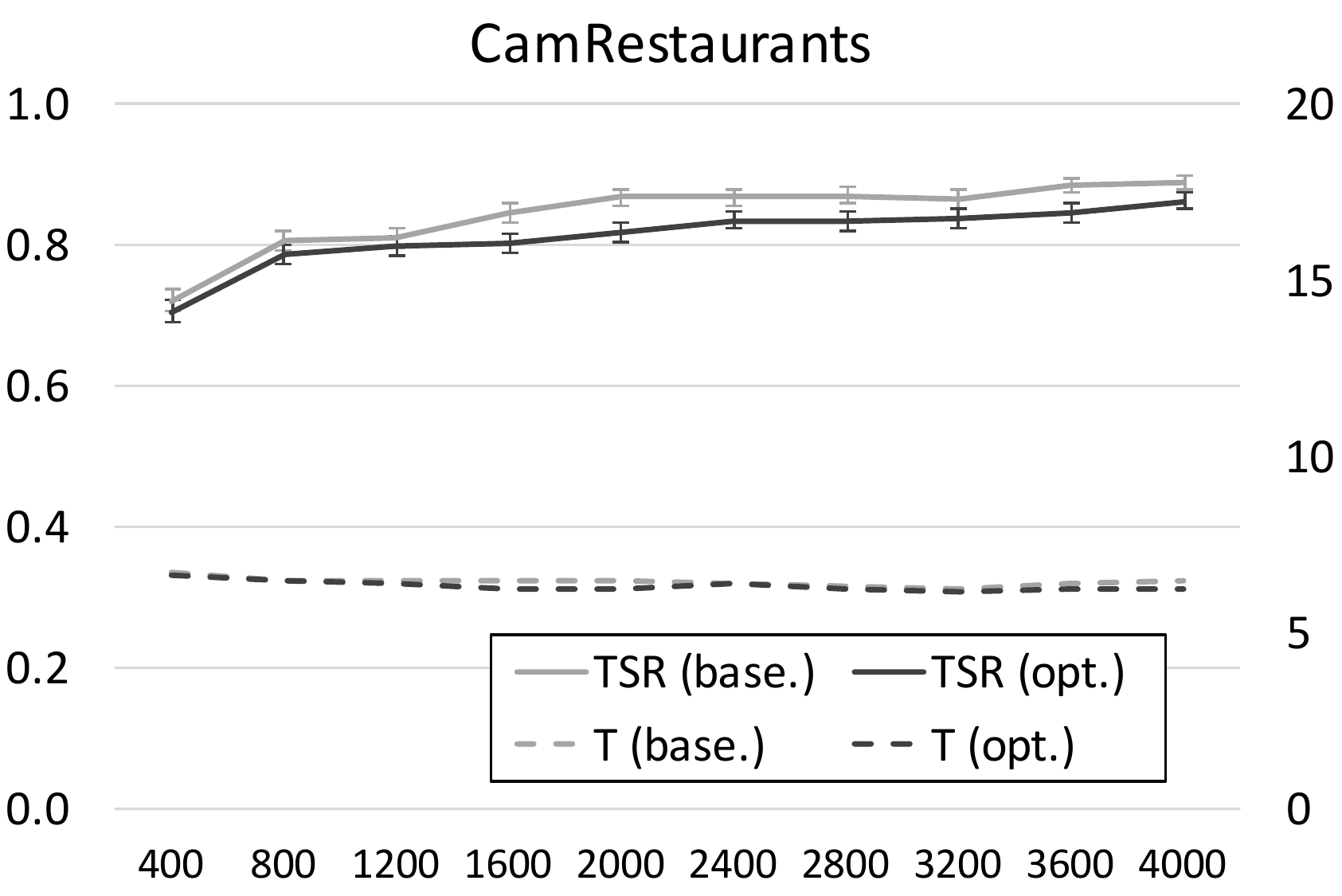}
  \end{minipage}%
  \hfill
  \begin{minipage}[b]{0.33\linewidth}
    \centering
    \includegraphics[width=\linewidth,height=\chartheight]{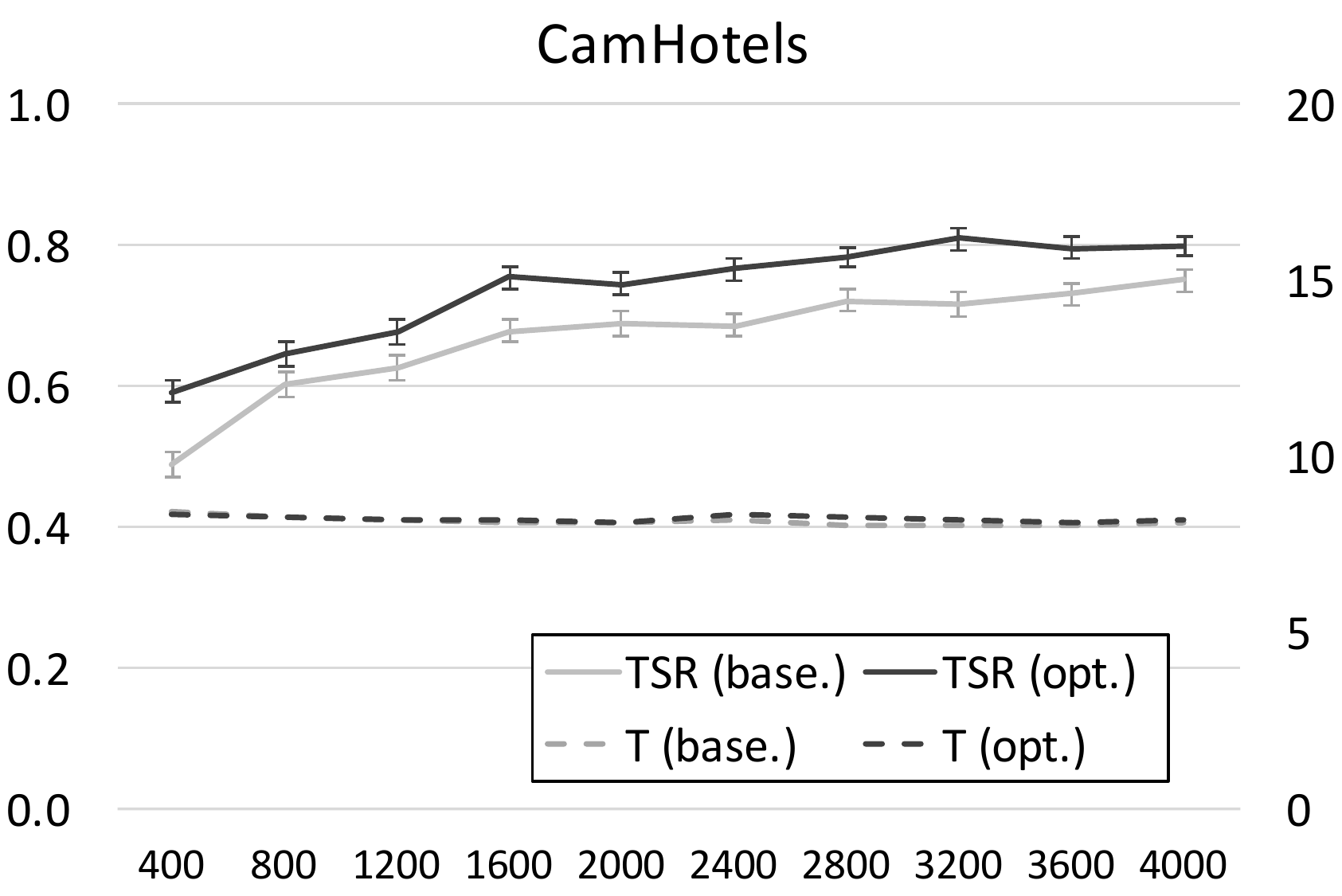}
  \end{minipage}%
  \hfill
    \begin{minipage}[b]{.33\linewidth}
    \centering
    \includegraphics[width=\linewidth,height=\chartheight]{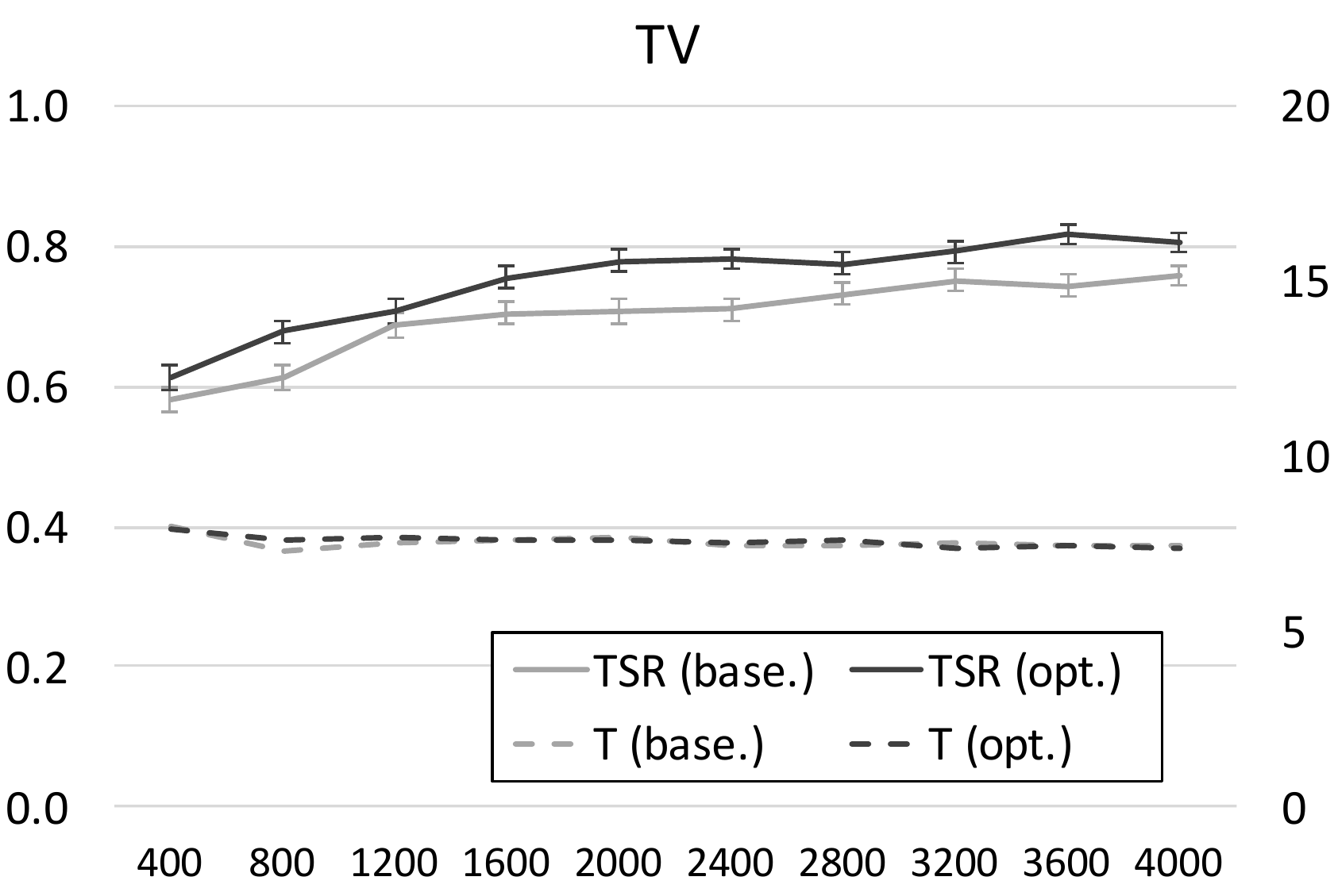}
  \end{minipage}%
  
  \begin{minipage}[b]{0.33\linewidth}
    \centering
    \includegraphics[width=\linewidth,height=\chartheight]{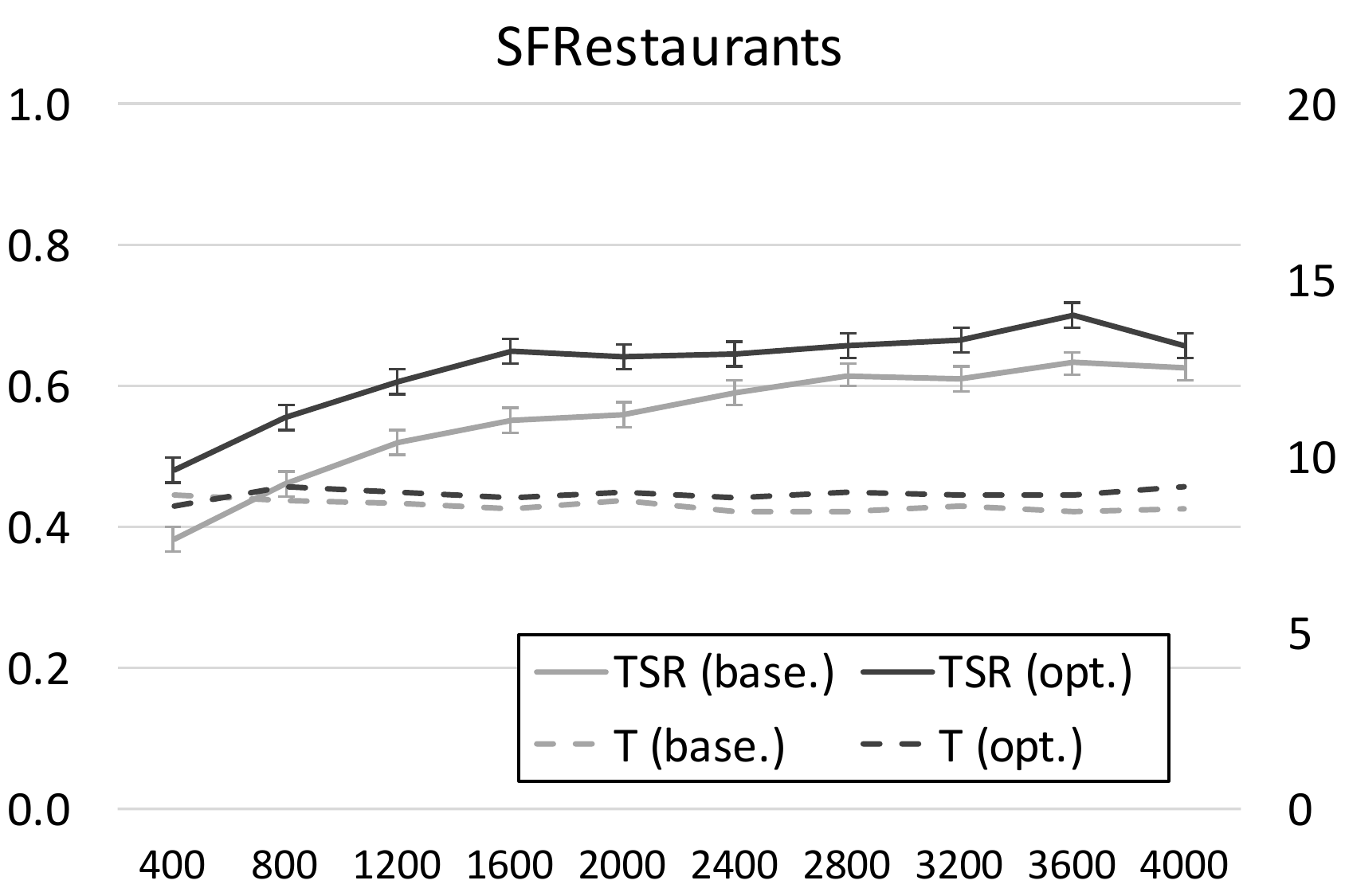}
  \end{minipage}%
  \hfill
  \begin{minipage}[b]{0.33\linewidth}
    \centering
    \includegraphics[width=\linewidth,height=\chartheight]{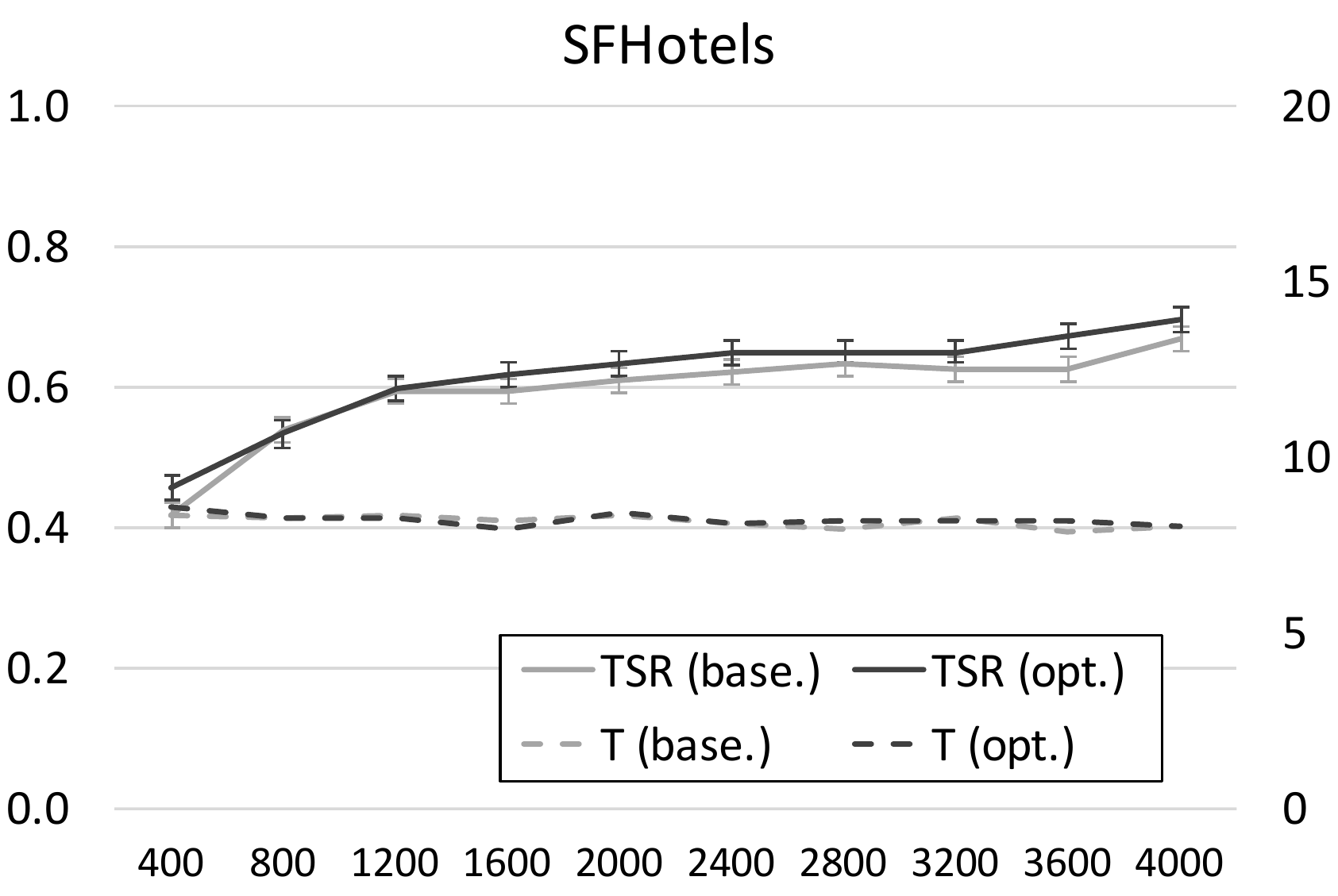}
  \end{minipage}%
  \hfill
    \begin{minipage}[b]{.33\linewidth}
    \centering
    \includegraphics[width=\linewidth,height=\chartheight]{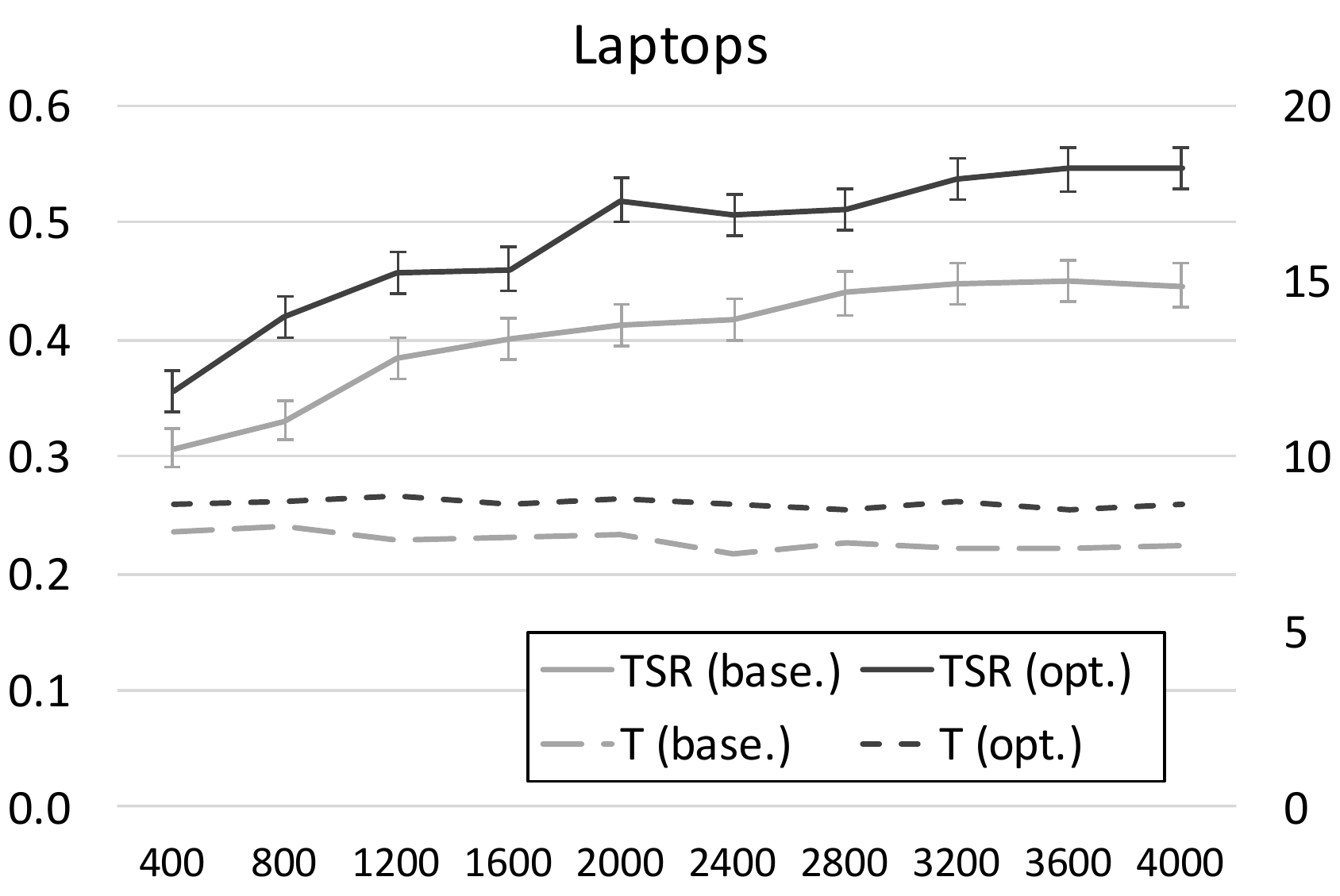}
  \end{minipage}%
  \caption{The task success rates (TSR, left axes) and dialogue length in number of turns (T, right axes) for all six domains comparing the baseline ($r_s^w = 20$, $w=(0.5,0.5)$) with the optimised balance. The horizontal axes show the number of training dialogues. Each data point is the average over five policies with different seeds where each policy is evaluated with 300 dialogues.  }
  \label{fig:result_curves}
\end{figure*}

The reward balancing method described in the previous section is applied to six domains: finding TVs, laptops, restaurants or hotels (the latter two in Cambridge and San Francisco). The following table depicts the domain statistics with the number of search constraints, the number of informational items the user can request, and the number of data-base entities: \vspace{0.1cm}

\noindent {\footnotesize \centering
\begin{tabular}{rccc}

    \toprule
    \textit{Domain} & \textit{\# constr.} & \textit{\# requests} & \textit{\# entities} \\
    \midrule
    CamRestaurants &  3  & 9    & 110 \\
    CamHotels & 5  &  11   & 33 \\
    SFRestaurants &  6  &  11   & 271 \\
    SFHotels &  6  &  10   & 182 \\
    TV & 6 & 14 & 94 \\
    Laptops &  11  &  21    & 126 \\
    \bottomrule
    \end{tabular}%
} \vspace{0.1cm}

For consistency with previous work~\cite{gasic2014gaussian,young2013,su2016acl} the rewards $r_s^w = 20$ and $r_l^w = -1$ are used representing the weight configuration $\mathbf{w} = (0.5,0.5)$. This results in $r_s = 40$ and $r_l = -2$. 

For the evaluation, simulated dialogues were created using the statistical spoken dialogue toolkit PyDial~\cite{ultes2017pydial}. It contains an agenda-based user simulator~\cite{schatzmann2009} with an error model to simulate the semantic error rate (SER) encountered in real systems due to the noisy speech channel.

A policy has been trained for each domain using multi-objective GPSARSA with 3,000 dialogues
and an SER of 15\%. Each policy was evaluated with 300 dialogues for each weight configuration in $\{(0.1,0.9),(0.2,0.8),\ldots,(0.9,0.1)\}$. The results in Figure~\ref{fig:weight_curves} are the averages of five trained policies with different random seeds. All curves follow a similar pattern: 
at some point, the success curve reaches a plateau where the performance does not increase any further with higher $w_s$.

The following weights were selected: CamRestaurants $w_s = 0.4$; CamHotels $w_s = 0.6$; SFRestaurants $w_s = 0.6$; SFHotels $w_s = 0.7$; TV $w_s = 0.6$; Laptops $w_s = 0.7$. These weights were selected by hand according to the success rate\footnote{Taking into account the overall performance and the proximity to the edge of the plateau. To compensate for possible inaccuracies of the MO-GPSARSA, the configuration right at the edge has not been chosen.} as well as the average dialogue length.


The selected weights were scaled
to keep the turn penalty $w_l^w$ constant at $-1$. Using these reward settings, each domain was evaluated with 4,000 dialogues in 10 batches. After each batch, the policies were evaluated with 300 dialogues. The final results shown in Table~\ref{tab:results} (selection of learning curves in Figure~\ref{fig:result_curves}) are compared to the baseline of $\mathbf{w} = (0.5,0.5)$ (i.e.~standard unoptimised reward component weight balance). 
Evidently, optimising the balance has a significant impact on the performance of the trained polices.

\begin{table}[t]
\setlength{\tabcolsep}{5pt}
  \centering
  \footnotesize
    \begin{tabular}{lccccc}
    \toprule
          & \multirow{2}[0]{*}{$r_{s}^w$} & \multicolumn{2}{c}{TSR} & \multicolumn{2}{c}{\# Turns} \\
          \cmidrule(l{2pt}r{2pt}){3-4} \cmidrule(l{2pt}r{2pt}){5-6}
          & & \textit{base.} &\textit{opt.} & \textit{base.} & \textit{opt.} \\
          \midrule
    CamRestaurants & 14    & 88.8\% & 86.2\% & 6.4   & 6.3 \\
    CamHotels & 30    & 75.1\% & 79.8\% & 8.1   & 8.2 \\
    SFRestaurants & 47    & 62.4\% & 65.7\% & 8.5   & 9.1 \\
    SFHotels & 30    & 66.7\% & 69.4\% & 8.0   & 8.0 \\
    TV    & 30    & 75.7\% & 80.5\% & 7.4   & 7.4 \\
    Laptops & 47    & 44.6\% & 54.6\% & 7.5   & 8.7 \\
    \bottomrule
    \end{tabular}%
  \caption{Task success rates (TSRs) and number of turns after 4,000 training dialogues using a success reward of 20 (baseline) compared to the optimised success reward $r_{s}^w$. All TSR differences are statistically significant ($t$-test, $p < 0.05$).}
  \label{tab:results}%
\end{table}%

To analyse the performance of multi-objective GPSARSA, policies were trained and evaluated for each reward balance with single-objective (SO) GPSARSA (see Figure~\ref{fig:weight_curves}) \textit{after} the weights had been selected. Each SO policy was trained with 1,000 dialogues and evaluated with 300 dialogues, all averaged over five runs. The success-weight curves for SORL clearly resemble the MORL curves for almost all domains except for CamRestaurants where it leads to an incorrect selection of weights. This may be attributed to the kernel used for multi-objective GPSARSA. 

It is worth noting that for the presented full MORL analysis, 3,000 training dialogues were necessary for each domain to find a good balance. This is significantly less than the 9,000 dialogues needed for the SORL analysis and this difference would increase further for  a finer grain search grid.

\section{Conclusion}
\label{sec:conclusion}

In this work, we have addressed the problem of finding a good balance between multiple rewards for learning dialogue policies. We have shown the relevance of the problem and demonstrated the usefulness of multi-objective reinforcement learning to facilitate the search for a suitable balance.  Using the proposed procedure, only one policy needs to be trained which can then be evaluated for an arbitrary number of reward balances thus drastically reducing the total amount of training dialogues needed.

We have proposed and employed an extension of the GPSARSA algorithm for multiple objectives and applied it to six domains. The experiments show the successful application of our method: the optimal balance improved task success without unduly impacting on dialogue length in all domains except CamRestaurants, where it is clear that the weight selection criteria failed.  In practice, this could have been easily trapped by applying a minimum weight to the success criteria. Furthermore, the domain-dependence of the reward balance has been confirmed.

For future work, the accuracy of the proposed multi-objective GPSARSA will be further improved with the ultimate goal of using the proposed method to directly learn a multi-objective policy through interaction with real users. To achieve this, alternative weight kernels will be explored. The resulting multi-objective policy may then directly be applied (without the need of re-training a single-objective policy) and the weights may even be adjusted according to a specific situation or user preferences.

Future work will also include an automatic method to find the optimal balance as well as investigating the relationship between the optimal success reward value and the domain characteristics (similar to \citet{papangelis2017}).



\section*{Acknowledgments}
Tsung-Hsien Wen, Pawe\l~Budzianowski and Stefan Ultes are supported
by  Toshiba  Research  Europe  Ltd,  Cambridge  Research Laboratory. This research was partly funded by the EPSRC grant EP/M018946/1 \textit{Open Domain Statistical Spoken Dialogue Systems}.

\section*{Data}
All experiments were run in simulation. The corresponding source code is included in the PyDial toolkit which can be found on www.pydial.org.

\bibliography{references}
\bibliographystyle{acl_natbib}

\end{document}